\documentclass{article}

\usepackage{PRIMEarxiv}

\usepackage[utf8]{inputenc} 
\usepackage[T1]{fontenc}    
\usepackage{hyperref}       
\usepackage{url}            
\usepackage{booktabs}       
\usepackage{amsfonts}       
\usepackage{nicefrac}       
\usepackage{microtype}      
\usepackage{lipsum}
\usepackage{graphicx}
\graphicspath{{media/}}     

\hypersetup{
    colorlinks=true,
    linkcolor=blue,
    filecolor=magenta,      
    urlcolor=cyan,
    pdftitle={Overleaf Example},
    pdfpagemode=FullScreen,
    }
\usepackage{graphicx}
\usepackage{amsmath}
\usepackage{amssymb}
\usepackage{booktabs}
\usepackage{float}
\usepackage{lipsum}
\usepackage{algorithm}
\usepackage{algpseudocode}
\usepackage[super]{nth}
\usepackage[symbol]{footmisc}

\title{Two-Step Active Learning for Instance Segmentation with Uncertainty and Diversity Sampling}
\author{Ke Yu\thanks{These authors contributed equally to this work}
  \textsuperscript{, }\thanks{Work completed during an internship at Google}\\
University of Pittsburgh\\
{\tt\small yu.ke@pitt.edu}
\And
Stephen Albro$^*$\\
Google\\
{\tt\small salbro@google.com}
\And 
Giulia DeSalvo\\
Google\\
{\tt\small giuliad@google.com}
\And 
Suraj Kothawade\\
Google\\
{\tt\small skothawade@google.com}
\And 
Abdullah Rashwan\\
Google\\
{\tt\small arashwan@google.com}
\And 
Sasan Tavakkol \\
Google\\
{\tt\small tavakkol@google.com}
\And 
Kayhan Batmanghelich \\
Boston University\\
{\tt\small batman@bu.edu}
\And 
Xiaoqi Yin \\
Google\\
{\tt\small yinx@google.com}
}

\begin{document}
\maketitle

\begin{abstract}
Training high-quality instance segmentation models requires an abundance of labeled images with instance masks and classifications, which is often expensive to procure. Active learning addresses this challenge by striving for optimum performance with minimal labeling cost by selecting the most informative and representative images for labeling. Despite its potential, active learning has been less explored in instance segmentation compared to other tasks like image classification, which require less labeling. In this study, we propose a post-hoc active learning algorithm that integrates uncertainty-based sampling with diversity-based sampling. Our proposed algorithm is not only simple and easy to implement, but it also delivers superior performance on various datasets. Its practical application is demonstrated on a real-world overhead imagery dataset, where it increases the labeling efficiency fivefold.
\end{abstract}

\section{Introduction}
Instance segmentation is the task of identifying and segmenting individual objects in an image, and it has a wide range of applications in real-world domains such as autonomous driving~\cite{de2017semantic, malbog2019mask}, medical imaging~\cite{chen2017dcan, johnson2020automatic}, and aerial imagery analysis~\cite{waqas2019isaid, champ2020instance}, among others. However, obtaining annotations for instance segmentation is considerably more expensive than other computer vision tasks due to the need for unique labeling of each instance and precise pixel-level segmentation for objects. This creates a significant bottleneck in the development and implementation of state-of-the-art  instance segmentation models.

Active learning is a technique designed to reduce labeling cost by iteratively selecting the most \emph{informative} samples from unlabeled data. There are two major ways to select the next batch to be labeled: uncertainty-based~\cite{lewis1994heterogeneous, lewis1995sequential, scheffer2001active, hwa2004sample, seung1992query}, which selects samples that model has low confidence in prediction, and diversity-based~\cite{nguyen2004active, xu2007incorporating, seneractive, van2012seeds, kading2016active},  which selects samples that are representative of the dataset. Both of these active learning approaches have been successfully applied in various computer vision tasks, including image classification~\cite{joshi2009multi, Li_2013_CVPR, Beluch_2018_CVPR}, object detection~\cite{li2021deep, choi2021active, kothawade2022talisman}, and semantic segmentation~\cite{yang2017suggestive, kuo2018cost, tan2019batch}. However, very few existing works have addressed active learning for instance segmentation, and the few existing works focus on uncertainty-based approaches.

Developing an effective active learning method for instance segmentation entails addressing several crucial factors. Instance segmentation models produce diverse types of output per instance, including class distribution, bounding box location, and a dense segmentation mask. This variety complicates the task of determining the most suitable uncertainty metric for active learning. Moreover, solely depending on the uncertainty metric can lead to redundancy, as most uncertain instances may share the same challenging semantic type (\textit{e.g.}, small pedestrians in background). Hence, incorporating diversity sampling to account for the semantic diversity within the unlabeled pool is crucial. In addition, information from individual instances needs to be aggregated into a score for image-level scoring and labeling, further complicating the active learning method's design. 

\begin{figure*}
\centering
    \includegraphics[width = 1.0 \textwidth]
    {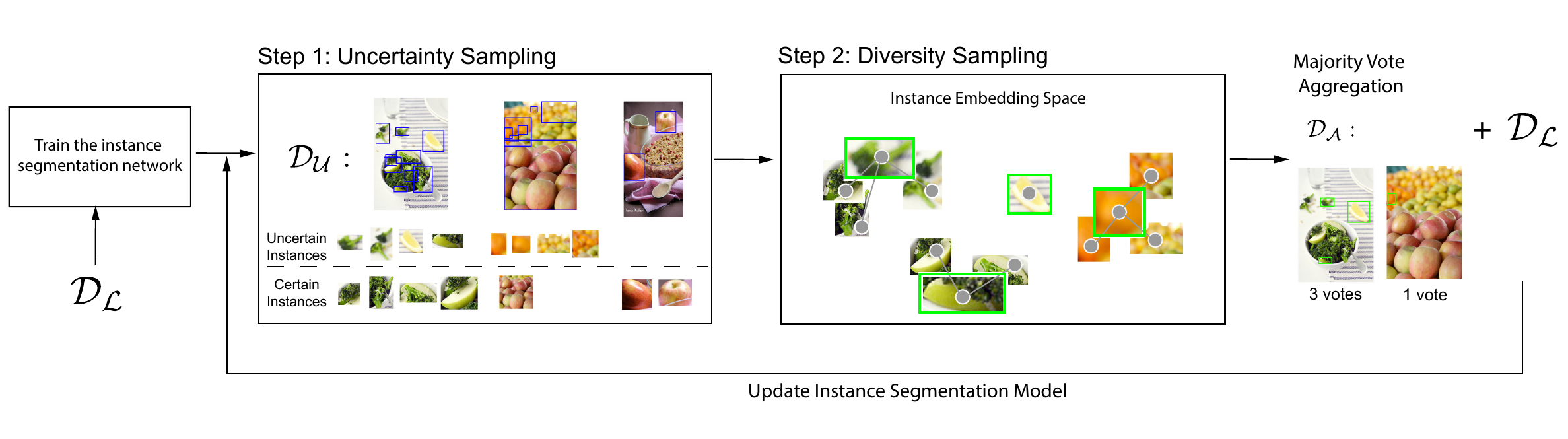}
    \caption{Schematic of TAUDIS. First, the uncertainty of each instance is assessed to identify the most informative instances from unlabeled data. Second, a graph-based maximum set cover algorithm is used to select the most representative instances among the uncertain ones. Finally, a majority vote approach selects images containing the most instances filtered from the previous steps for labeling. Symbols $\mathcal{D_L, D_U, D_A}$ represent labeled, unlabeled, and annotated set, respectively. 
    }
    \label{fig:main}
\end{figure*}

We propose TAUDIS, a \textbf{T}wo-step \textbf{A}ctive learning algorithm that combines \textbf{U}ncertainty and \textbf{D}iversity sampling strategies for \textbf{I}nstance \textbf{S}egmentation. In the first step, TAUDIS uses uncertainty sampling to identify an initial set of the most informative object instances. In the second step, we extract region feature maps from an intermediate convolutional layer in the network to represent each instance and apply the \emph{maximum set cover} algorithm~\cite{bateni2018optimal} to select a diverse subset of instances from the most uncertain instances identified eariler. Finally, to select images for labeling, we use a majority vote approach that prioritizes images containing the most informative instances selected by the two-step algorithm until the labeling budget is met. Fig.~\ref{fig:main} provides an overview of the proposed method. We evaluated TAUDIS on two instance segmentation datasets: MS-COCO~\cite{lin2014microsoft}, and a proprietary overhead imagery dataset. The results show that TAUDIS consistently outperforms the compared active learning strategies.

\section{Related Work}
Uncertainty-based methods select informative samples based on their ambiguities. 
In image classification, uncertainty is assessed via softmax probabilities using metrics such as least confidence~\cite{wang2016cost}, margin~\cite{joshi2009multi}, and entropy~\cite{ranganathan2017deep}. For semantic segmentation, pixel-wise class probability maps are used to construct image-level~\cite{kuo2018cost, tan2019batch} or region-level~\cite{siddiqui2020viewal, casanova2020reinforced} uncertainty scores. In object detection, uncertainty scores are first calculated for each bounding box, then combined to the image level, with previous research exploring the aggregation of scores across multiple locations~\cite{brust2018active, li2021deep} or multiple losses~\cite{kao2019localization, yoo2019learning}. However, finding the optimal uncertainty metric for instance segmentation is challenging due to the choices of representing uncertainties as either classification or segmentation uncertainties at the instance level, along with multiple options for aggregating them to the image-level. In this paper, we explore numerous combinations of uncertainty metrics and aggregation methods in the context of instance segmentation. 

Diversity-based active learning methods aim to identify samples that are representative of data distribution, with methods like Core-set~\cite{sener2017active}, VAAL~\cite{sinha2019variational}, CDAL~\cite{agarwal2020contextual}, and DBAL~\cite{gal2017deep} being developed for deep neural networks. The fusion of uncertainty and diversity sampling, demonstrated in approaches by Elhamifar \textit{et al.}~\cite{elhamifar2013convex}, Yin \textit{et al.}\cite{yin2021self}, Yang \textit{et al.}~\cite{yang2017suggestive} , and Wu \textit{et al.}~\cite{wu2022entropy}, becomes increasingly populate in active learning due to the combined benefits. To our knowledge, our method is the first to integrate both sampling strategies in active learning for instance segmentation.

Despite the higher annotation cost, research on active learning for instance segmentation has been limited. Wang \textit{et al.}~\cite{wang2020semi} proposed a triplet uncertainty metric combining model predictions from classes, bounding boxes, and segmentation masks, but this approach fails to consider diversity and requires modifications to the Mask R-CNN~\cite{he2017mask} architecture. Our method, on the other hand, recognizes diversity to eliminate redundancy and can be integrated with any existing architecture, including more recent transformer based models like Mask2Former~\cite{cheng2021mask2former}. Recently, Tang \textit{et al.}~\cite{tang2022active} proposed an active learning method trained with point supervision, yet it assumes pre-existing class labels and bounding boxes, which is not feasible for raw real-world images. In contrast, our method, which makes image-level sampling decisions without prior annotations, is more broadly applicable.

\section{Method}

\subsection{Model Training}
\label{sec:model_training}
We initiate the process of active learning by training an instance segmentation model, denoted as $\mathcal{M}_\theta$, on a labeled set of data $\mathcal{D_L}$. Our method is compatible with any instance segmentation model utilizing region-level features for mask prediction. In this study, we employ Mask R-CNN~\cite{he2017mask} as our model, given its extensive use in instance segmentation applications. During model inference, $\mathcal{M}_\theta$ generates $N$ detected instances $\{t_n\}_{n=1}^{N}$ for a given input image $x$, where $n$ represents instance index. For each instance $t_n$, we obtain its associated class-probability distribution $p_n$, sigmoid masks $m_n$, and instance embedding $r_n$. The instance embedding $r_n$ is computed as the average pooling of the regional feature map, which can be extracted from intermediate layers in either the object classification or object segmentation branch. For our experiments, we choose the feature map of the last convolutional layer in the object segmentation branch.

\subsection{Instance-level Uncertainty Measures}
\label{sec:uncertainty_sampling}
The first step of our method involves identifying uncertain instances from unlabeled images $\mathcal{D_U}$ based on the current state of $\mathcal{M}_\theta$. Unlike active learning methods for image classification that rely on image-level uncertainty scores, our approach uses instance-level uncertainty measures to select the top candidates. We explore three metrics to measure uncertainty at the instance-level: classification margin ($\text{CM}_n$), classification entropy ($\text{CE}_n$) and segmentation entropy ($\text{SE}_n$). Definitions of these metrics can be found in Appendix~\ref{app:instance_uncertainty_metric}.

\subsection{Instance-level Diversity Sampling}
\label{sec:diversity_sampling}
Uncertainty-based sampling alone may not yield satisfactory results when the sampled batch contains a considerable amount of redundancy~\cite{wei2015submodularity, kothawade2021similar}. This challenge is particularly relevant in instance segmentation, where the instance-level uncertainty scores are influenced not only by the semantic properties of objects but also by factors such as their size and spatial arrangement (\textit{et al.} small pedestrians in the background). To avoid redundancy, our method oversamples uncertain instances beyond the designated budget in the first step, and subsequently selects a diverse subset in the second step.

We formulate the diversity sampling as a \emph{maximum k-set cover problem}. Formally, let $\mathcal{T_F}$ denote all detected instances in $\mathcal{D_U}$ and $\mathcal{T_C}$ denote the subset of most uncertain instances in $\mathcal{T_F}$. Our objective is to select a subset of instances $\mathcal{T_D}$, $\mathcal{T_D} \subset \mathcal{T_C}$, that is highly representative of $\mathcal{T_F}$. To achieve this we build an undirected similarity graph, $G(V, E)$, such that vertices, $V$, are the instances in $\mathcal{T_F}$ and edges, $E$, represent the similarity between instances. To quantify the similarity, $s_{i,j}$, between instances $t_i, t_j \in \mathcal{T_F}$, we use the cosine similarity between their respective embeddings $r_{i}$ and $r_{j}$. The edges are kept only if $s_{i,j}$ is larger than a similarity threshold, $\sigma$. To limit our samples to $\mathcal{T_C}$, we select a subset of vertices, $V_C=\{v_i \mid v_i \in {T_C}\}$, and all associated edges, $E_C=\{e_{i,j} \mid v_i \in {V_C} \text{ or } v_j \in {V_C}\}$. Note that vertices in a graph defined as $G(V_C, E_C)$, may have dangling edges where one end of the edge is in $\mathcal{T_C}$ but the other end is in $\mathcal{T_F-T_C}$.
We define our maximum k-set cover problem by the bipartite graph $G(V_C, U_C, E_C)$, where $U_C = \{u_i | e_{i,j} \in E_C \text{ or } e_{j,i} \in E_C\}$. $V_C$ is the subset to sample from and $U_C$ is the universe we want to cover. We determine the optimal subset of instances $\mathcal{T_D}$ by maximizing coverage of $G(V_C, U_C, E_C)$ constrained to ${|\mathcal{T_D}|} = k$, using the distributed submodular optimization algorithm introduced in ~\cite{bateni2018optimal}.

We use two hyperparameters, $\alpha$ and $\beta$ ($\alpha > \beta > 1$), as multipliers to the image-level annotation budget $\mathcal{B}$ to adjust the upsampling and downsampling of instances in the first and second steps, respectively. In particular, we first select $|\mathcal{T_C}|=\alpha \mathcal{B}$ most uncertain instances and then refine this set to the $|\mathcal{T_D}|=\beta \mathcal{B}$ most representative instances. 

\subsection{Majority Vote Aggregation}
\label{sec:instance_aggregation}
To acquire annotations for the selected instances, we prioritize images with the highest number of instances in $\mathcal{T_D}$. This is based on the intuition that images containing a large number of uncertain and diverse instances likely encompass important visual concepts that the model has yet to learn. We compute the number of instances $n_D$ in $\mathcal{T_D}$ for each image in the unlabeled pool $\mathcal{D_U}$ and rank the images by $n_D$ in descending order. The top $\mathcal{B}$ images are then chosen for annotation. Subsequently, the $\mathcal{B}$ newly annotated images $\mathcal{D_A}$ are removed from the unlabeled dataset and added to the labeled dataset, which is utilized to retrain the instance segmentation model. The complete algorithm for our proposed active learning method is outlined in Algorithm~\ref{alg:alfis}.

\begin{algorithm}
\caption{TAUDIS (Illustration in Fig.~\ref{fig:main})}\label{alg:alfis}
\begin{algorithmic}[1]
\Require Labeled data: $\mathcal{D_L}$, Unlabeled data: $\mathcal{D_U}$, Model: $\mathcal{M}_\theta$, Budget: $\mathcal{B}$, Number of rounds: $\mathcal{I}$, Hyperparameters: $\alpha$, $\beta$ $(\alpha > \beta > 1)$, $0<\sigma<1$
\For{$i$ = 1 : $\mathcal{I}$}
\State Train $\mathcal{M}_{\theta}$ on $\mathcal{D_L}$\vspace{1px}
\State Obtain detected instances $ \mathcal{T_F} \leftarrow
\mathcal{M}_\theta(\mathcal{D_U})$\vspace{2px}
\State Compute uncertainty scores for each instance in $\mathcal{T_F}$\vspace{2px}
\State \parbox[t]{\dimexpr\linewidth-\algorithmicindent}{Select top uncertain instances $\mathcal{T_C} \subset \mathcal{T_F}$. $|\mathcal{T_C}|=\alpha \mathcal{B}$}\vspace{2px}
\State \parbox[t]{\dimexpr\linewidth-\algorithmicindent}{Construct $\mathcal{S} \in \mathbb{R}^{|\mathcal{T_C}| \times |\mathcal{T_F}|}$, the pairwise similarity matrix between $\mathcal{T_C}$ and $\mathcal{T_F}$ instance embeddings. Matrix elements smaller than $\sigma$ are set to zero.}\vspace{4px}
\State \parbox[t]{\dimexpr\linewidth-\algorithmicindent}{Determine top representative instances $\mathcal{T_D} \subset \mathcal{T_C}$ via maximum k-set cover on a bipartite graph defined by $\mathcal{S}$ and $k=|\mathcal{T_D}|=\beta \mathcal{B}$. Rows of $\mathcal{S}$ define the subsets, and columns define the universe of the maximum k-set cover problem.}\vspace{2px}
\State Select $\mathcal{B}$ images with the most instances in $\mathcal{T_D}$\vspace{2px}
\State Annotate selected images $\mathcal{D_A}$. $|\mathcal{D_A}| = \mathcal{B}$\vspace{2px}
\State $\mathcal{D_U} := \mathcal{D_U} - \mathcal{D_A}$, $\mathcal{D_L} := \mathcal{D_L} + \mathcal{D_A}$\vspace{2px}
\EndFor
\State \Return $\mathcal{M}_\theta$
\end{algorithmic}
\end{algorithm}

\section{Experiments}
\subsection{Experimental Setup}
\textbf{Datasets.} Our method is evaluated on two datasets: COCO~\cite{lin2014microsoft} and OVERHEAD, a proprietary building segmentation dataset. COCO comprises around 118k training images and 5k validation images. The OVERHEAD dataset consists of overhead images with bounding boxes and per-instance segmentation masks for buildings. It contains around 50k training images and 6k validation images. For each dataset, we use the training set as the unlabeled data and evaluate the trained models on the validation set.

\textbf{Active Learning Setting.} For our experiments, we begin by randomly selecting around 25\% of the images from the unlabeled set to form the initial labeled data. In each subsequent active learning cycle, we add a small, fixed batch size of images to the labeled set. The active learning iterations continue until at least 90\% of the unlabeled samples have been selected for labeling. Based on the results of the ablation study (Appendix~\ref{app:ablation_stuyd}), we choose segmentation entropy $\text{SE}_n$ as the instance-level uncertainty metric in our TAUDIS method. For further implementation details, please refer to Appendix~\ref{app:implementation_details}.

\textbf{Baselines.}
We compare our method against several baselines, including random sampling, uncertainty-based sampling using metrics such as classification margin, classification entropy, and segmentation entropy, as well as diversity-based sampling methods such as Core-set~\cite{sener2017active} and Round-robin~\cite{citovsky2021batch}. We also compare our method to a variation of TAUDIS that combine uncertainty and diversity sampling at the image level. Additional details about the baselines can be found in Appendix~\ref{app:baselines}. For evaluation, we use the mean average precision at 50\% intersection over union (mAP50) as the performance metric for assessing the instance segmentation models.

\subsection{Results}
\begin{figure*}[ht]
\centering
    \includegraphics[width = 1.0 \textwidth]
    {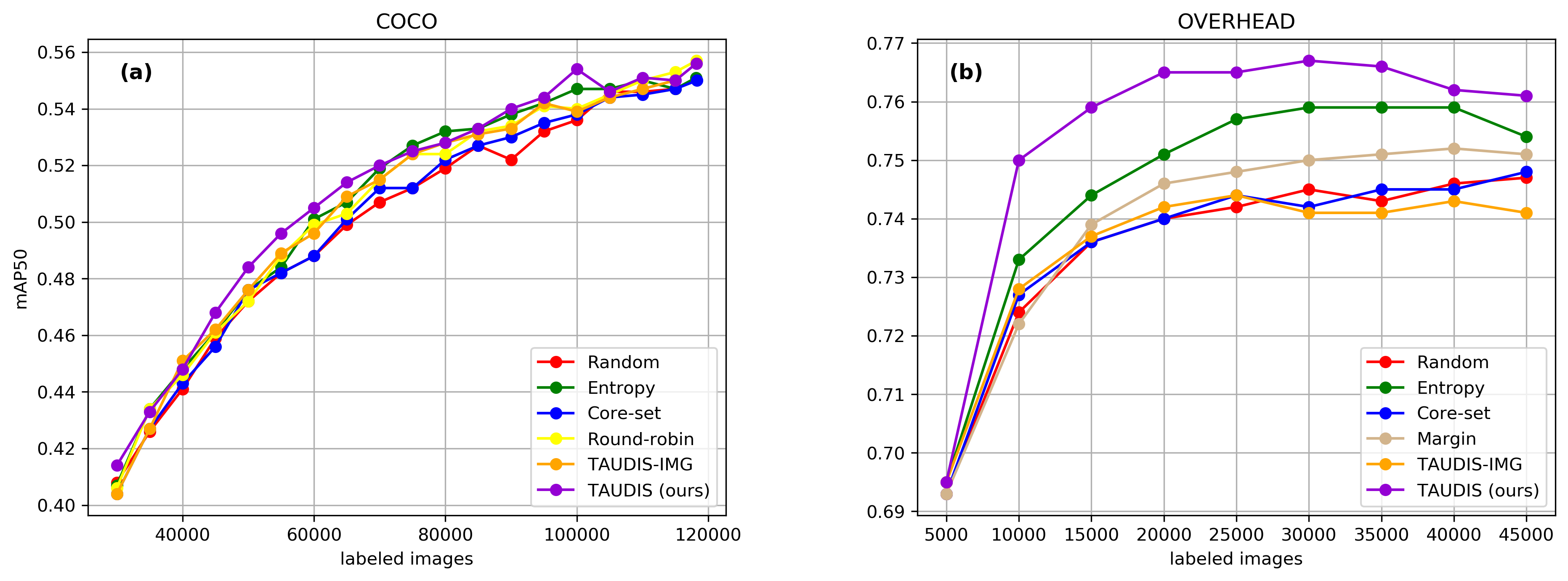}
    \caption{Results for instance segmentation.}
    \label{fig:main_results}
\end{figure*}

\noindent \textbf{COCO.} Fig.~\ref{fig:main_results}(a)
compares TAUDIS with various baseline methods: random, uncertainty sampling with segmentation entropy, and two diversity-based methods (\textit{i.e.,} Coreset and Round robin). TAUDIS consistently outperforms other baselines particularly during the early activation cycles. The result also highlight the effectiveness of segmentation entropy as a metric for uncertainty in instance segmentation, as it is the second-best performer. Furthermore, the superiority of TAUDIS over TAUDIS-IMG and Core-set suggests that instance-level features are a better proxy for diversity than image-level features, for the sub-image task of instance segmentation.\\

\noindent \textbf{OVERHEAD.} Fig.~\ref{fig:main_results}(b) shows the results on our building segmentation data set. Since there is only one class (buildings), we omitted the round-robin baseline. Notably, TAUDIS achieves superior performance using only 10k images compared to the random strategy's performance with 45k images, resulting in an improvement in labeling efficiency of nearly \emph{fivefold}. This finding highlights the effectiveness of the diversity-based sampling strategy, even in a single-class setting.

\section{Conclusion}
In conclusion, our study presents a novel active learning algorithm for instance segmentation, addressing the lack of post-hoc methods in this field. By combining uncertainty and diversity sampling at the instance level, our algorithm outperforms various baselines across multiple datasets. Moreover, we demonstrate its practical value in a real-world satellite imagery dataset, achieving an \emph{fivefold} improvement in labeling efficiency. As future work, we plan to explore the use of transformer-based segmentation architectures, such as MaskFormer~\cite{cheng2021maskformer} and Mask2Former~\cite{cheng2021mask2former}, which take a unified view of instance and semantic segmentation. We expect that using these richer embeddings may synergize effectively with our diversity-based active learning strategy, potentially unlocking an efficient batch active learning approach for panoptic segmentation.

\bibliographystyle{unsrt}
\bibliography{reference}

\clearpage
\appendix
\setcounter{table}{0}
\setcounter{figure}{0}
\renewcommand\thefigure{\thesection.\arabic{figure}}    
\renewcommand\thetable{\thesection.\arabic{table}}  

\section{Instance-level Uncertainty Metrics}
Definition of classification margin ($\text{CM}_n$), classification entropy ($\text{CE}_n$) and segmentation entropy ($\text{SE}_n$):
\label{app:instance_uncertainty_metric}
\begin{equation}
\label{eq:cm}
    {\text{CM}_n} = p_n(c_1) - p_n(c_2),
\end{equation}
\begin{equation}
\label{eq:ce}
    {\text{CE}_n} = - \sum_{k=1}^{K} p_n(c_k) \log p_n(c_k),
\end{equation}
\begin{equation}
\label{eq:se}
    {\text{SE}_n} = \frac{1}{WH} \sum_{w, h} \text{BCE} \big(m_n^*(w,h)\big),
\end{equation}
where $p_n(c_k)$ denotes the predicted probability of class $k$, and $c_1, c_2$ denote the \nth{1} and \nth{2} most probable class labels under $\mathcal{M}_\theta$, $m_n^* = m_n(c_1)$ is the binary mask of the winning class, $W, H$ are the width and height of this binary mask. $\text{BCE}$ refers to the binary cross-entropy function.

\section{Ablation Study}
\label{app:ablation_stuyd}
\begin{figure}[h]
\centering
    \includegraphics[width = 0.45 \textwidth]
    {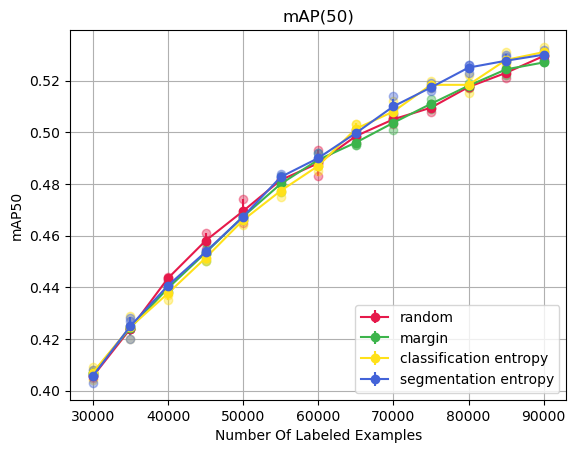}
    \caption{Ablation study: Comparing uncertainty metrics for instance segmentation on COCO dataset. Instance-level uncertainty metrics are computed and then aggregated to the image-level using simple averaging. Sampling with entropy-based metrics outperforms random strategy after 60k examples, while sampling using classification margin does not yield better results than random strategy. 
    }
    \label{fig:ue_coco_5k_s30k_3tr_trial_mAP50}
\end{figure}

\section{Implementation Details}
\label{app:implementation_details}
To ensure a fair comparison, all active learning methods in our experiments utilize the same instance segmentation network, namely Mask R-CNN~\cite{he2017mask} with FPN~\cite{lin2017feature} backbone. Batch size, learning rate, and other hyperparameters are set on a per data set basis, according to what works well for that data set. All hyperparameters are kept constant across a data set during experimentation. For each active learning cycle, we reset the model parameters and conduct training for 25k steps, employing Stochastic Gradient Descent (SGD) optimizer with momentum value set at 0.997. A mini-batch size of 128 is used for COCO
. The initial learning rate is set to 0.12 (0.012 for OVERHEAD) and is reduced by a factor of 10 at the 15k and 20k step marks.

The experiments on each dataset begin with an identical, randomly-selected set of annotated images from the unlabeled data. We set the hyperparameter $\alpha$ to be 2.5x higher than the average number of instances per image in the seed set; 150 for COCO and 50 for OVERHEAD. 
We set $\beta$ to be lower than $\alpha$, 1-2x higher than the same number of instances; 40 for COCO and 30 for OVERHEAD. 
We set the cosine similarity threshold $\sigma = 0.8$ for all data sets.

\section{Baselines}
\label{app:baselines}

\noindent \textbf{Average Classification Margin.} This sampling strategy assigns the average instance classification margin (Eq.~\ref{eq:cm}) as the uncertainty value for each image and selects the ones with lowest values for annotation.

\noindent \textbf{Weighted Classification Entropy.} This sampling strategy assigns the instance-size-weighted classification entropy as the uncertainty value for each image and selects the ones with highest values for annotation. The weighted classification entropy is defined as:
\begin{equation}
    \text{WCE} = \sum_{n=1}^N s_n \times \text {CE}_n,
\end{equation}
where $s_n$ is the ratio of the size of instance $t_n$ to the overall image size, and $\text{CE}_n$ is the instance-level classification entropy defined in Eq.~\ref{eq:ce}.

\noindent \textbf{Weighted Segmentation Entropy.} This sampling strategy assigns the instance-size-weighted segmentation entropy as the uncertainty value for each image and selects the ones with highest values for annotation. The weighted segmentation entropy is defined as:
\begin{equation}
\label{eq:wse}
    \text{WSE} = \sum_{n=1}^N s_n \times \text {SE}_n,
\end{equation}
where $s_n$ is the ratio of the size of instance $t_n$ to the overall image size, and $\text{SE}_n$ is the instance-level segmentation entropy defined in Eq.~\ref{eq:se}.

\noindent \textbf{Core-Set.} We use the Core-set~\cite{sener2017active} as a diversity-based sampling baseline. The objective of this method is to find a
set of points, such that distance of any data point from its nearest cover-point is minimized. We use the features of the bottom layer of Feature Pyramid Network (FPN)~\cite{lin2017feature} in Mask R-CNN as the image representation and Euclidean distance as the similarity metric.

\noindent \textbf{Round-Robin.} We define the image-level, class-specific uncertainty metric $\text{WSE}(c_k)$ as the weighted segmentation entropy of instances whose winning class is $c_k$. Following ~\cite{citovsky2021batch}, we then sample images with the highest $\text{WSE}(c_k)$ from each class $c_k$ in a round-robin style until reaching the budget limit $\mathcal{B}$.

\noindent \textbf{TAUDIS-IMG} To compare which aggregation level is more effective, instance or image, we have developed another two-step active learning algorithm, named TAUDIS-IMG, that employs uncertainty metrics and diversity sampling at the image level. Differing from its instance-level counterpart, TAUDIS-IMG uses image-level uncertainty measure (e.g., WSE) to select most informative samples at the first step. The diversity sampling is formulated as a maximum k-cover problem on a bipartite graph, where vertices are \textit{image-level} embeddings obtained from the FPN, and edges denote the cosine similarity between these embeddings. We use features from the bottom layer of the FPN in Mask R-CNN as the image embedding. The full details of the algorithm is presented in Algorithm~\ref{alg:alfis-img}.

\begin{algorithm}
\caption{TAUDIS-IMG}\label{alg:alfis-img}
\begin{algorithmic}[1]
\Require Labeled data: $\mathcal{D_L}$, Unlabeled data: $\mathcal{D_U}$, Model: $\mathcal{M}_\theta$, Budget: $\mathcal{B}$, Number of rounds: $\mathcal{I}$, Hyperparameters: $\alpha$, $0<\sigma<1$
\For{$i$ = 1 : $\mathcal{I}$}
\State Train $\mathcal{M}_{\theta}$ on $\mathcal{D_L}$\vspace{1px}
\State Compute $\text{WSE}$ (Eq.~\ref{eq:wse}) for all the images in $\mathcal{D_U}$\vspace{1px}
\State \parbox[t]{\dimexpr\linewidth-\algorithmicindent}{Select top uncertain images  $\mathcal{D_C} \subset \mathcal{D_U}$ with highest WSE values.  $|\mathcal{D_C}|=\alpha \mathcal{B}$}\vspace{1px}
\State \parbox[t]{\dimexpr\linewidth-\algorithmicindent}{Construct $\mathcal{S} \in \mathbb{R}^{|\mathcal{D_C}| \times |\mathcal{D_F}|}$, the pairwise similarity matrix between $\mathcal{D_C}$ and $\mathcal{D_F}$ image embeddings. Matrix elements smaller than $\sigma$ are set to zero.} \vspace{1px}
\State \parbox[t]{\dimexpr\linewidth-\algorithmicindent}{Determine top representative image $\mathcal{D_A} \subset \mathcal{D_C}$ via maximum k-set cover on a bipartite graph defined by $\mathcal{S}$ and $k=\mathcal{B}$. Rows of $\mathcal{S}$ define the subsets, and columns define the universe of the maximum k-set cover problem.}\vspace{1px}
\State Annotate selected images $\mathcal{D_A}$. $|\mathcal{D_A}| = \mathcal{B}$\vspace{1px}
\State $\mathcal{D_U} := \mathcal{D_U} - \mathcal{D_A}$, $\mathcal{D_L} := \mathcal{D_L} + \mathcal{D_A}$\vspace{1px}
\EndFor
\State \Return $\mathcal{M}_\theta$
\end{algorithmic}
\end{algorithm}

\clearpage
\onecolumn
\section{Qualitative Results}
\begin{figure*}[h]
\centering
    \includegraphics[width = 1.0 \textwidth]
    {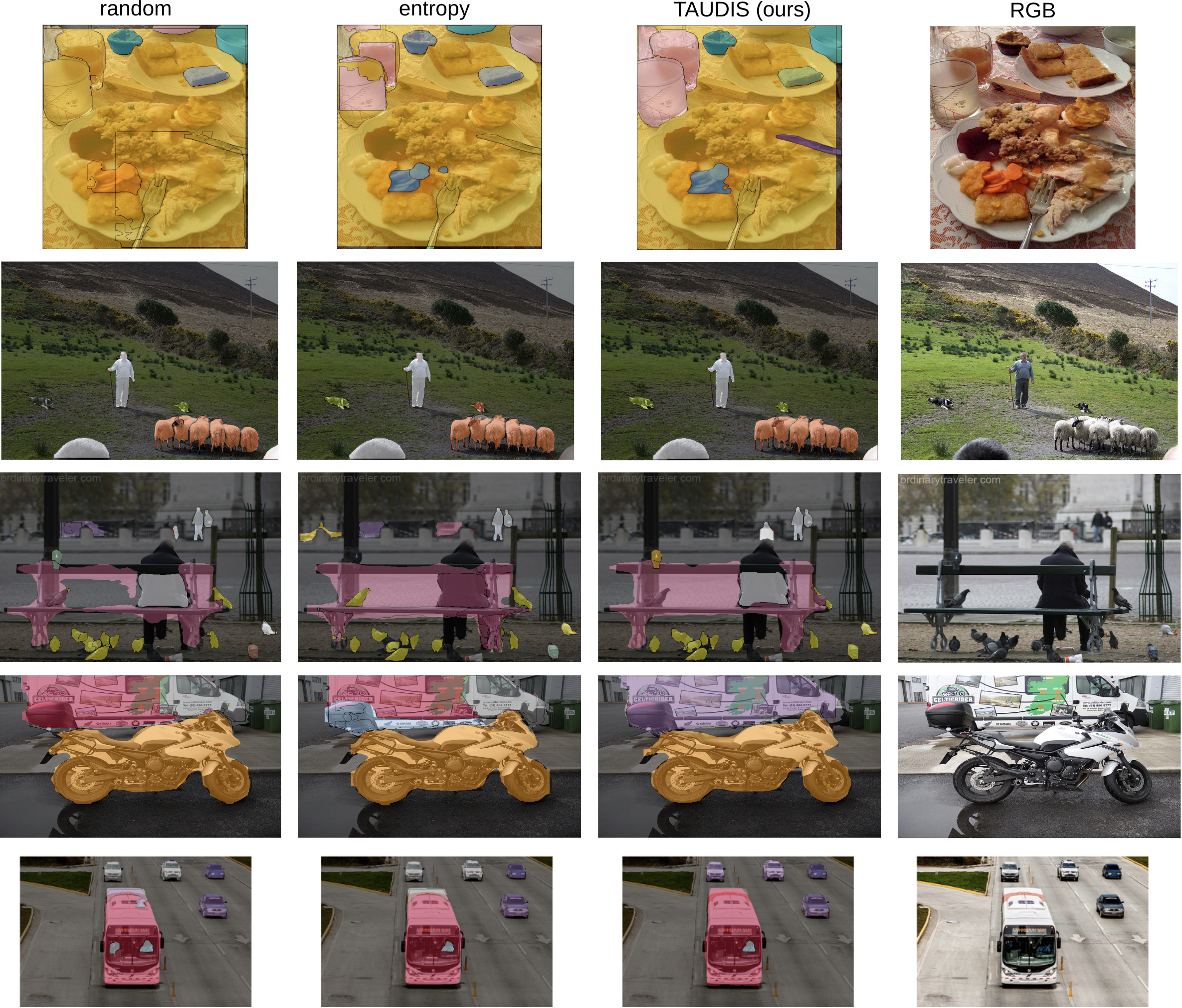}
    \caption{Results on COCO, showing model performance at the \nth{6} sampling iteration for random, entropy, and TAUDIS sampling strategies. Instances are outlined and examples of the same class are shown in the same color.}
    \label{fig:coco_qualitative}
\end{figure*}

\end{document}